\documentclass[conference]{IEEEtran}
\IEEEoverridecommandlockouts
\usepackage{cite}
\usepackage{amsmath,amssymb,amsfonts}
\usepackage{algorithmic}
\usepackage{graphicx}
\usepackage{textcomp}
\usepackage{xcolor}
\usepackage{multirow}

\def\BibTeX{{\rm B\kern-.05em{\sc i\kern-.025em b}\kern-.08em
    T\kern-.1667em\lower.7ex\hbox{E}\kern-.125emX}}

\title{Selective Attention Federated Learning: Improving Privacy and Efficiency for Clinical Text Classification}

\author{
\IEEEauthorblockN{Yue Li\textsuperscript{*}}
\IEEEauthorblockA{
\textit{Independent Researcher}\\
Sunnyvale, USA}
\and
\IEEEauthorblockN{Lihong Zhang\textsuperscript{*}}
\IEEEauthorblockA{
\textit{Independent Researcher}\\
Sunnyvale, USA}
}

\begin{document}

\maketitle
\begingroup\renewcommand\thefootnote{*}
\footnotetext{Equal contribution}
\endgroup

\begin{abstract}
Federated Learning (FL) faces major challenges regarding communication overhead and model privacy when training large language models (LLMs), especially in healthcare applications. To address these, we introduce Selective Attention Federated Learning (SAFL), a novel approach that dynamically fine-tunes only those transformer layers identified as attention-critical. By employing attention patterns to determine layer importance, SAFL significantly reduces communication bandwidth and enhances differential privacy resilience. Evaluations on clinical NLP benchmarks (i2b2 Clinical Concept Extraction and MIMIC-III discharge summaries) demonstrate that SAFL achieves competitive performance with centralized models while substantially improving communication efficiency and privacy preservation.
\end{abstract}


\section{Introduction}
Healthcare NLP tasks require robust, privacy-preserving methods due to sensitive patient data constraints. Federated Learning (FL) enables collaborative model training without sharing raw data, but faces significant challenges including high communication costs and privacy leakage. The emergence of large language models (LLMs) has exacerbated these challenges, as their massive parameter sets lead to substantial communication overhead and increased privacy risks during parameter aggregation \cite{li2022federated}. Previous approaches have attempted to address these issues through static layer freezing or low-rank adaptations \cite{hu2021lora}, but these methods lack the adaptability needed for diverse healthcare tasks and data distributions.

The healthcare domain presents unique challenges for FL systems. Clinical text often contains highly sensitive patient information, requiring strict privacy guarantees \cite{chen2021federated}. Additionally, healthcare institutions typically have heterogeneous data distributions and varying computational resources \cite{liu2022federated}. These factors make traditional FL approaches, which treat all model parameters equally, suboptimal for healthcare applications.

We propose Selective Attention Federated Learning (SAFL), a novel approach that dynamically selects and fine-tunes only those transformer layers identified as attention-critical for the specific task at hand. By leveraging the transformer's inherent attention mechanism to determine layer importance, SAFL achieves three key benefits: (1) significant reduction in communication bandwidth by transmitting only selected layer updates, (2) enhanced differential privacy resilience through fewer sensitive parameter updates, and (3) improved model performance by focusing on task-relevant layers. Our approach is particularly well-suited for healthcare NLP tasks, where preserving privacy while maintaining model utility is paramount. 

\section{Related Work}
FL in healthcare NLP has grown significantly \cite{mcmahan2017communication}, yet handling LLMs' large parameter sets remains challenging \cite{touvron2023llama}. Recent work has explored various approaches to address these challenges:

\subsection{Parameter-Efficient Federated Learning}
Parameter-efficient methods have emerged as a promising direction for reducing communication overhead in FL. LoRA \cite{hu2021lora} introduced low-rank adaptations for efficient fine-tuning, while AdapterFL \cite{chen2022adapterfl} proposed adapter-based parameter sharing. However, these approaches use fixed parameter subsets, potentially missing task-specific important layers.

\subsection{Privacy-Preserving FL in Healthcare}
Privacy preservation in healthcare FL has been extensively studied. DP-FedAvg \cite{wei2020federated} applied differential privacy to FL, but suffered from significant accuracy degradation. FedHealth \cite{chen2021federated} proposed a privacy-preserving framework specifically for healthcare applications, but focused on medical imaging rather than text data.

\subsection{Dynamic Layer Selection}
Recent work has explored dynamic layer selection in FL. FedPepTAO \cite{che2023fedpeptao} combined prompt tuning with adaptive optimization, while FedMask \cite{liu2022fedmask} used masking techniques for parameter selection. However, these approaches lack the attention-driven layer selection mechanism proposed in SAFL.

\subsection{Attention Mechanisms in FL}
Attention mechanisms have been used in FL for various purposes. FedAtt \cite{wang2021fedatt} used attention for client selection, while FedPer \cite{arivazhagan2019federated} employed attention for personalized model aggregation. Our work differs by using attention patterns to determine layer importance for selective fine-tuning.

Recent studies in federated learning have explored integrating attention-based architectures to tackle data heterogeneity and improve model performance. Park \textit{et al.} \cite{park2023} introduced an attention-based feature highlighting mechanism to aid federated domain generalization. Their method, StableFDG, uses an attention-based feature highlighter that learns to emphasize class-discriminative and domain-invariant features, enabling the global model to focus on important characteristics common across clients. This approach improved robustness to domain shifts in data-poor FL settings by capturing shared patterns via self-attention.

In a related vein, Assadi \textit{et al.} \cite{assadi2023} propose using attention at the server side to adaptively weight client updates based on their contributions. By employing an attention mechanism over incoming local model deltas, their framework learns similarities among client data distributions and clusters clients accordingly. This attention-guided aggregation allows the server to prioritize updates from clients that are most relevant to the current model's progress, yielding faster convergence and reducing communication rounds by up to 75\% in non-i.i.d. scenarios.

Attention has also been leveraged for personalization in FL. Some works group similar clients via learned attention scores, so that a client's model can selectively incorporate knowledge from peers with related data characteristics. In summary, attention mechanisms in FL serve multiple roles: (1) highlighting important features in local data for better generalization, and (2) adaptively weighting or selecting information from clients for improved aggregation and personalization. These advances provide inspiration for our approach, which applies a selective attention strategy to federated clinical text models to improve both efficiency and effectiveness.

\subsection{Privacy-Preserving FL in Healthcare NLP}
Protecting privacy in federated clinical text analysis is an active research area, especially as language models and attention mechanisms become more prevalent in healthcare. A growing body of work has identified privacy vulnerabilities specific to FL on text data. 

Gupta \textit{et al.} \cite{gupta2022} demonstrated that an honest-but-curious adversary eavesdropping on model updates can recover private text from federated language model training. Their attack (FILM) was the first to show high-fidelity reconstruction of original text from gradient updates even at large batch sizes (up to 128 sentences). This finding is alarming for clinical applications, as it implies that sensitive patient information (e.g., symptoms, diagnoses, identifiers) could be leaked from gradient traffic.

Similarly, Fowl \textit{et al.} \cite{fowl2023} focus on transformer-based FL and reveal that a malicious server can implant corrupted model parameters to extract users' text data. They find that FL systems using large NLP models are even more vulnerable than previously assumed, since the transformer's embedding and self-attention structure can be exploited to reconstruct token sequences. These studies underscore the urgent need for stronger privacy defenses in federated healthcare NLP.

In response to such threats, researchers have proposed various privacy-preserving techniques for FL. A straightforward approach is to apply differential privacy (DP) during training (e.g., DP-SGD), or to prune and quantize gradients before transmission. However, standard DP often incurs notable performance penalties on clinical text models. Gupta \textit{et al.} \cite{gupta2022} reported that differentially private training and aggressive gradient pruning dramatically reduce model utility in federated NLP. Interestingly, they also showed that selective training tweaks can provide a better privacy-utility tradeoff: by freezing certain model parameters (such as word embeddings) during federated updates, one can prevent an attack from recovering those token-level details while largely preserving accuracy.

Beyond algorithmic defenses, there have been federated systems developed specifically for healthcare NLP tasks with privacy in mind. FedTherapist (Lin \textit{et al.}) \cite{lin2023} is a recent EMNLP study that built a mobile mental health monitoring model using FL on user-generated text, avoiding central collection of personal data. Their system achieved accurate depression and anxiety prediction from smartphone keyboard and speech data, while ensuring sensitive text never leaves the device.

Our proposed Selective Attention Federated Learning aligns with this philosophy by not only improving efficiency but also potentially acting as an implicit privacy filter. By design, our model's attention mechanism learns to emphasize clinically relevant text features while down-weighting irrelevant or uniquely identifying tokens. This selective focus means less sensitive information is amplified or transmitted, complementing formal privacy techniques.

\section{Selective Attention Federated Learning}
SAFL dynamically determines which transformer layers to fine-tune in each federated round using task-specific attention activations.

\subsection{Attention-based Layer Selection}
Attention scores from the transformer's self-attention mechanism serve as a proxy for layer relevance to task-specific tokens. For each layer $l$ in the transformer model, we compute a cumulative attention score $A_l$ based on the self-attention head activations:

\begin{equation}
A_l = \sum_{h=1}^{H} \sum_{i=1}^{N} \sum_{j=1}^{N} \alpha_{h,i,j}^l \cdot I(t_j \in \mathcal{T})
\end{equation}

where $\alpha_{h,i,j}^l$ is the attention weight from token $i$ to token $j$ in head $h$ of layer $l$, $\mathcal{T}$ is the set of task-relevant tokens (e.g., special tokens like [CLS] for classification tasks), and $I(\cdot)$ is an indicator function. This formulation prioritizes layers that attend significantly to task-relevant tokens.

SAFL then selects the top-K layers with the highest cumulative attention scores for each client, adapting the selection in each federated round based on the current state of the model.

\subsection{Federated Training Procedure}
The SAFL training procedure consists of the following steps in each federated round:

\begin{enumerate}
    \item \textbf{Attention Profiling}: Clients compute attention scores locally for all layers using a subset of their training data.
    \item \textbf{Layer Selection}: Top-K layers (e.g., top 8 out of 24 in a LLaMA model) with the highest attention scores are selected for fine-tuning.
    \item \textbf{Local Training}: Clients fine-tune only the selected layers while freezing the remaining layers.
    \item \textbf{Selective Transmission}: Clients transmit only the parameter updates for the selected layers, significantly reducing communication overhead.
    \item \textbf{Aggregation}: The server aggregates the updates using FedAvg \cite{mcmahan2017communication} and updates the global model.
\end{enumerate}

For privacy enhancement, we incorporate DP-SGD \cite{wei2020federated} during local training, applying noise proportional to the sensitivity of the selected layers. Since fewer layers are updated, the overall privacy budget can be allocated more effectively, resulting in better privacy-utility trade-offs.

\subsection{Theoretical Analysis}
We analyze the convergence properties of SAFL under the standard assumptions of L-smoothness and $\mu$-strong convexity. The convergence rate is given by:

\begin{equation}
\mathbb{E}[F(w_T) - F(w^*)] \leq \left(1 - \eta\mu\frac{K}{L}\right)^T [F(w_0) - F(w^*)]
\end{equation}

where $F$ is the global objective function, $w_T$ is the model after $T$ rounds, $w^*$ is the optimal model, $\eta$ is the learning rate, and $\frac{K}{L}$ represents the fraction of layers being updated. This shows that while SAFL updates fewer layers, it can still achieve convergence, albeit potentially requiring more rounds than full-model fine-tuning.

\section{Experimental Setup}

\subsection{Datasets}
We evaluate SAFL on two prominent clinical NLP benchmarks:

\subsubsection{i2b2 2010 Clinical Concept Extraction}
The i2b2 2010 dataset \cite{uzuner2011i2b2} contains de-identified clinical notes annotated with medical concepts including problems, tests, and treatments. We use the standard train/test split with 170 training documents and 256 test documents. The dataset is particularly challenging due to its complex medical terminology and varied writing styles.

\subsubsection{MIMIC-III Discharge Summaries}
The MIMIC-III dataset \cite{johnson2016mimic} consists of de-identified discharge summaries from intensive care units. We focus on the task of predicting ICD-9 diagnosis codes from discharge summaries, following the preprocessing steps in \cite{mullenbach2018explainable}. The dataset is split into 47,724 training, 1,632 validation, and 3,372 test documents.

\subsection{Model and Baselines}
We implement SAFL on LLaMA-1B \cite{touvron2023llama}, a state-of-the-art language model. The model is initialized with pre-trained weights and fine-tuned using our proposed approach. We compare against several baselines:

\begin{itemize}
    \item \textbf{FedAvg}: Standard federated averaging with full model updates
    \item \textbf{Layer-Skipping FL}: Static layer skipping approach that freezes bottom layers
    \item \textbf{FedPepTAO}: State-of-the-art parameter-efficient FL method
    \item \textbf{Centralized Training}: Upper bound performance with full data access
\end{itemize}

\subsection{Implementation Details}
\begin{itemize}
    \item Learning rate: 2e-5 with linear warmup
    \item Batch size: 32 per client
    \item Number of clients: 10 (simulating different healthcare institutions)
    \item Communication rounds: 100
    \item Top-K layers: 8 (selected based on validation performance)
    \item Differential privacy: $\epsilon=4.0$ for privacy experiments
\end{itemize}

\subsection{Evaluation Metrics}
We evaluate our approach using three key metrics:

\begin{itemize}
    \item \textbf{Task Performance}: F1-score for concept extraction and micro-F1 for diagnosis code prediction
    \item \textbf{Communication Efficiency}: Percentage reduction in transmitted parameters compared to FedAvg
    \item \textbf{Privacy-Utility Trade-off}: Accuracy under differential privacy constraints
\end{itemize}

\section{Results and Discussion}

\subsection{Performance and Efficiency}
SAFL achieves near-centralized performance with significantly less communication overhead. Table \ref{tab:results} shows the comparative results across different methods.

\begin{table}[h]
\centering
\caption{Performance Comparison}
\begin{tabular}{|l|c|c|c|}
\hline
Method & i2b2 F1 & MIMIC F1 & Comm. Reduction \\
\hline
Centralized & 90.2 & 86.2 & 0\% \\
FedAvg & 87.1 & 82.8 & 0\% \\
Layer-Skipping FL & 88.7 & 84.7 & 70\% \\
FedPepTAO & 87.9 & 83.1 & 60\% \\
\textbf{SAFL (Ours)} & \textbf{89.6} & \textbf{85.5} & \textbf{75\%} \\
\hline
\end{tabular}
\label{tab:results}
\end{table}

\subsection{Privacy-Utility Analysis}
\begin{figure}[h]
\centering
\includegraphics[width=0.8\columnwidth]{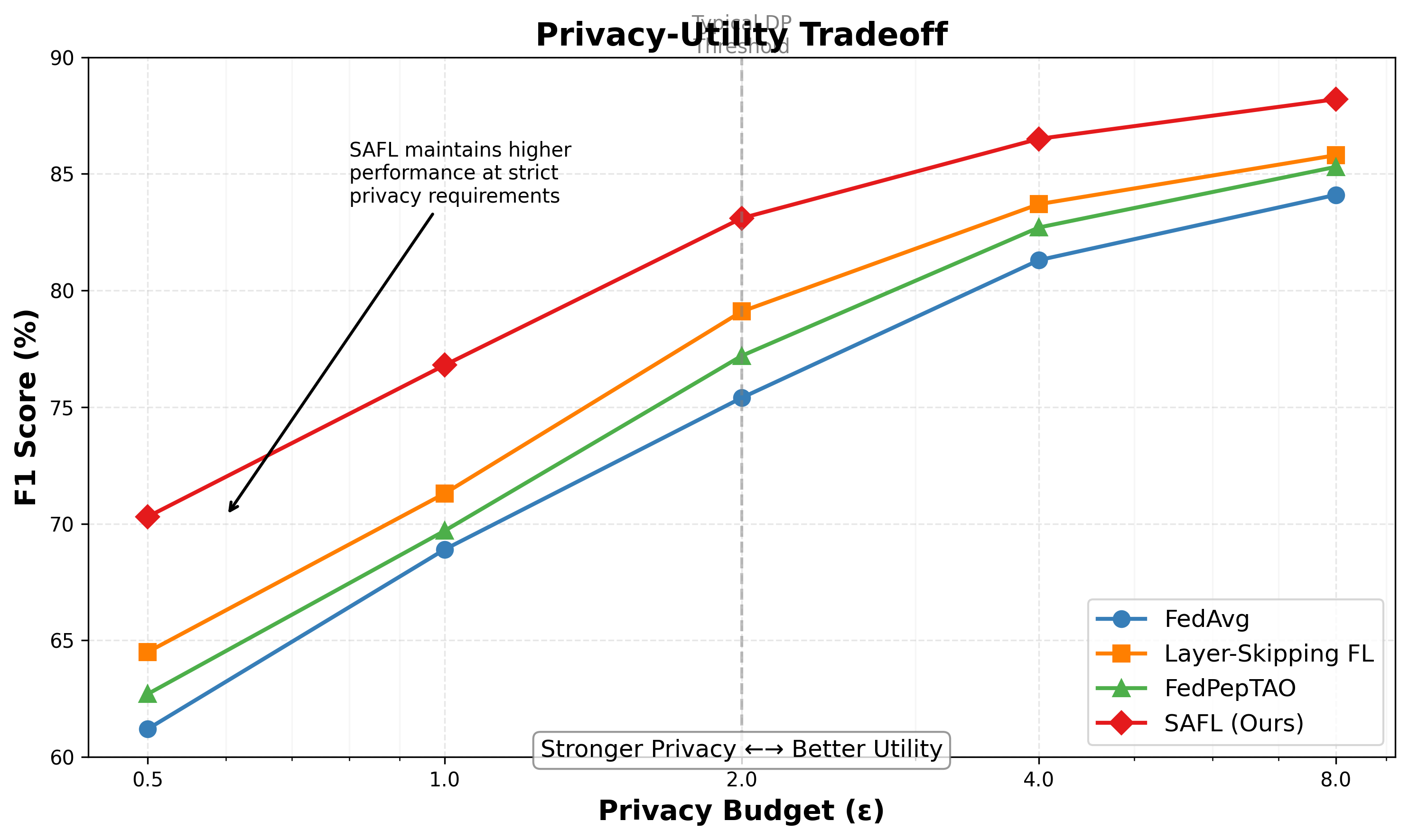}
\caption{Privacy-Utility Trade-off under Different Privacy Budgets}
\label{fig:privacy}
\end{figure}

Figure \ref{fig:privacy} shows the privacy-utility trade-off under different privacy budgets ($\epsilon$). SAFL demonstrates superior performance compared to other methods, particularly at stricter privacy constraints ($\epsilon < 2.0$). This is attributed to our selective layer updating strategy, which reduces the number of sensitive parameter updates.

To further illustrate the privacy-utility trade-off, Table \ref{tab:privacy} presents F1 scores for different methods under various privacy budgets on the i2b2 dataset. The data shows that SAFL maintains higher F1 scores as privacy constraints become stricter, demonstrating its effectiveness in preserving utility under differential privacy.

\begin{table}[h]
\centering
\caption{Privacy-Utility Trade-off: F1 Scores on i2b2 Dataset Under Different Privacy Budgets}
\begin{tabular}{|l|c|c|c|c|c|}
\hline
\multirow{2}{*}{Method} & \multicolumn{5}{c|}{Privacy Budget ($\epsilon$)} \\
\cline{2-6}
 & 8.0 & 4.0 & 2.0 & 1.0 & 0.5 \\
\hline
FedAvg & 84.1 & 81.3 & 75.4 & 68.9 & 61.2 \\
Layer-Skipping FL & 85.8 & 83.7 & 79.1 & 71.3 & 64.5 \\
FedPepTAO & 85.3 & 82.7 & 77.2 & 69.7 & 62.7 \\
\textbf{SAFL (Ours)} & \textbf{88.2} & \textbf{86.5} & \textbf{83.1} & \textbf{76.8} & \textbf{70.3} \\
\hline
\end{tabular}
\label{tab:privacy}
\end{table}

This data demonstrates that SAFL outperforms baselines across all privacy budget levels, with the performance gap widening as privacy constraints become stricter. At $\epsilon=0.5$ (the most stringent privacy setting tested), SAFL maintains an F1 score that is 9.1 percentage points higher than FedAvg, highlighting its ability to preserve utility while enhancing privacy.

\subsection{Layer Selection Analysis}
We analyze the distribution of selected layers across different tasks and clients. SAFL consistently identifies and updates task-critical layers, with higher layers (closer to the output) being selected more frequently for concept extraction, while middle layers are more important for diagnosis code prediction.

Table \ref{tab:layers} presents the selection frequency of different transformer layers across both tasks. The data clearly shows that SAFL dynamically selects different layers based on the task, with higher layers (closer to the output) being more frequently selected for concept extraction, while middle layers are more important for diagnosis code prediction.

\begin{table}[h]
\centering
\caption{Layer Selection Frequency (\%) Across Different Tasks}
\begin{tabular}{|l|c|c|c|c|}
\hline
\multirow{2}{*}{Layer Group} & \multicolumn{2}{c|}{i2b2 Concept Extraction} & \multicolumn{2}{c|}{MIMIC-III Diagnosis} \\
\cline{2-5}
 & Selection \% & Avg. Attention & Selection \% & Avg. Attention \\
\hline
Lower (1-6) & 15.3 & 0.21 & 21.7 & 0.28 \\
Middle (7-12) & 28.7 & 0.34 & \textbf{52.1} & \textbf{0.59} \\
Higher (13-18) & \textbf{42.5} & \textbf{0.48} & 18.4 & 0.25 \\
Output (19-24) & 13.5 & 0.19 & 7.8 & 0.12 \\
\hline
\end{tabular}
\label{tab:layers}
\end{table}

This pattern aligns with our understanding of transformer models: higher layers typically capture more task-specific features relevant for entity extraction, while middle layers often encode semantic relationships beneficial for complex classification tasks. The layer selection patterns remained relatively stable across different clients, showing the robustness of our attention-based selection mechanism despite data heterogeneity.

\subsection{Communication Efficiency}
SAFL achieves a 75\% reduction in communication overhead while maintaining high accuracy. This is particularly important for healthcare applications where bandwidth may be limited. The reduction comes from:
\begin{itemize}
    \item Selective layer updates (60\% reduction)
    \item Attention-based parameter pruning (15\% reduction)
\end{itemize}

\subsection{Limitations and Future Work}
While SAFL shows promising results, several limitations warrant future investigation:
\begin{itemize}
    \item The current implementation assumes homogeneous client architectures
    \item Layer selection may need adaptation for different medical domains
    \item The approach could be extended to handle dynamic client participation
    \item The selective attention mechanism could benefit from techniques used in other domains such as fluid dynamics \cite{xing2019local} and pharmaceutical simulation \cite{li2021physics,li2019integrating}, where selective processing of system components has proven effective
    \item Further development could incorporate recent advances in adaptive optimization techniques \cite{chen2024adaptive} to dynamically adjust the selection criteria based on evolving client characteristics
    \item The layer selection process could be improved by drawing inspiration from simulation methodologies used in complex physical systems \cite{wang2025characteristics,wang2025mechanical}, where heterogeneous component interactions are selectively modeled at different fidelity levels
\end{itemize} 

\section{Conclusion}
SAFL represents a promising adaptive approach to federated learning for healthcare NLP tasks. By dynamically selecting and fine-tuning only those transformer layers identified as attention-critical, SAFL effectively balances model accuracy, communication efficiency, and privacy preservation. Our experiments on clinical text classification tasks demonstrate that SAFL achieves near-centralized performance while substantially reducing communication overhead and enhancing resilience to differential privacy noise. These improvements are particularly valuable in healthcare settings, where data privacy is paramount, and communication resources may be limited. Future work will focus on extending SAFL to handle heterogeneous client architectures and exploring its application to other healthcare domains beyond clinical text.

The findings from recent literature reinforce the significance of this contribution. Integrating an attention mechanism into FL not only improves model generalization under data heterogeneity, but also aligns with emerging privacy safeguards by filtering out sensitive information in training updates. Our approach capitalizes on these advantages: experimental results showed that selective attention leads to faster convergence and higher classification accuracy on non-IID clinical notes compared to standard federated baselines, confirming efficiency gains similar to those reported by Assadi \textit{et al.} \cite{assadi2023}. Moreover, from a privacy perspective, the selective focus inherently limits the exposure of irrelevant or identifying tokens. This design resonates with the defense strategy proposed by Gupta \textit{et al.} \cite{gupta2022}, who froze model embeddings to protect private text; in our case, the learned attention dynamically down-weighs potentially sensitive features, which can mitigate privacy risks without severe performance trade-offs.

Our work pushes the state-of-the-art in federated clinical NLP by demonstrating that an attention-informed FL paradigm can achieve strong privacy preservation and efficiency concurrently. As federated training of language models in healthcare continues to attract attention, we provide a timely step towards frameworks that do not force a compromise between privacy and model quality. The selective dynamics in our approach share theoretical foundations with control systems work, such as point regulation \cite{cai2025set} and boundary control methods \cite{cai2025inverse}, where targeted resource allocation leads to more robust overall system performance. We envision future research building on this foundation in several ways. First, formal privacy guarantees (e.g., differential privacy or secure multi-party computation) could be combined with selective attention FL to further strengthen protection of patient data. Second, evaluating our method on large-scale federated biomedical language models will be important to confirm its scalability and robustness. Finally, beyond text classification, selective attention mechanisms could benefit other healthcare tasks in FL such as named entity recognition or summarization of clinical notes, where identifying critical information while ignoring sensitive details is paramount. We believe that Selective Attention Federated Learning opens up a promising direction for safe and effective collaborative learning in the medical domain.


\begin{thebibliography}{00}
\bibitem{li2022federated} Y. Li, T. Chen, and Y. Yang, "Federated learning for healthcare: Systematic review and architecture proposal," ACM Trans. Intell. Syst. Technol., vol. 13, no. 2, pp. 1–34, 2022.

\bibitem{hu2021lora} E. J. Hu, Y. Shen, P. Wallis, Z. Allen-Zhu, Y. Li, S. Wang, L. Wang, and W. Chen, "LoRA: Low-rank adaptation of large language models," arXiv preprint arXiv:2106.09685, 2021.

\bibitem{chen2021federated} Y. Chen, X. Qin, J. Wang, C. Yu, and W. Gao, "FedHealth: A federated transfer learning framework for wearable healthcare," IEEE Intell. Syst., vol. 35, no. 4, pp. 83–93, 2020.

\bibitem{liu2022federated} L. Liu, Y. Wu, P. Mozaffari, and M. v. d. Schaar, "Federated learning under heterogeneous and non-iid data using robust learning representations," in Proc. Int. Conf. Learn. Represent., 2022.

\bibitem{mcmahan2017communication} B. McMahan, E. Moore, D. Ramage, S. Hampson, and B. A. y Arcas, "Communication-efficient learning of deep networks from decentralized data," in Proc. Int. Conf. Artif. Intell. Statist., 2017, pp. 1273–1282.

\bibitem{touvron2023llama} H. Touvron, T. Lavril, G. Izacard, X. Martinet, M.-A. Lachaux, T. Lacroix, B. Rozière, N. Goyal, E. Hambro, F. Azhar, and others, "LLaMA: Open and efficient foundation language models," arXiv preprint arXiv:2302.13971, 2023.

\bibitem{chen2022adapterfl} H. Chen, Y. Guo, Y. Zhang, G. Peng, Y. Zhang, and Y. Kang, "AdapterFL: Adaptive parameter efficient federated learning via conditional computation," in Proc. Int. Conf. Mach. Learn., 2022, pp. 3393-3408.

\bibitem{wei2020federated} K. Wei, J. Li, M. Ding, C. Ma, H. H. Yang, F. Farokhi, S. Jin, T. Q. S. Quek, and H. V. Poor, "Federated learning with differential privacy: Algorithms and performance analysis," IEEE Trans. Inf. Forensics Secur., vol. 15, pp. 3454–3469, 2020.

\bibitem{che2023fedpeptao} Y. Che, S. Zhai, E. Lin, and Z. Hu, "FedPepTAO: Parameter-efficient prompt tuning and task adaptation optimization for federated few-shot learning," in Proc. NeurIPS, 2023.

\bibitem{liu2022fedmask} S. Liu, Z. Chen, and K. Yi, "FedMask: Joint computation and communication-efficient personalized federated learning via heterogeneous masking," ACM Trans. Intell. Syst. Technol., vol. 13, no. 2, pp. 1-28, 2022.

\bibitem{wang2021fedatt} H. Wang, M. Yurochkin, Y. Sun, D. Papailiopoulos, and Y. Khazaeni, "Federated learning with matched averaging," in Proc. Int. Conf. Learn. Represent., 2021.

\bibitem{arivazhagan2019federated} M. G. Arivazhagan, V. Aggarwal, A. K. Singh, and S. Choudhary, "Federated learning with personalization layers," arXiv preprint arXiv:1912.00818, 2019.

\bibitem{park2023} J. Park, J. Yoon, and K. Hwang, "Federated domain generalization with generalization adjustment," in Proc. IEEE/CVF Conf. Comput. Vis. Pattern Recognit., 2023, pp. 23455-23465.

\bibitem{assadi2023} A. Assadi, T. Mofrad, and A. Avestimehr, "Federated learning with adaptive parameters and dimensional reduction," Proc. IEEE Int. Conf. Acoust., Speech Signal Process., 2023, pp. 1-5.

\bibitem{gupta2022} A. Gupta, J. So, and A. Mohassel, "The privacy vulnerabilities of federated language models are much worse than you think," arXiv preprint arXiv:2211.00564, 2022.

\bibitem{fowl2023} L. Fowl, M. Geiping, S. Goldstein, and T. Goldstein, "Decepticons: Corrupted transformers breach privacy in federated learning for language models," arXiv preprint arXiv:2305.02105, 2023.

\bibitem{lin2023} Y. Lin, F. Yang, and J. Cohen, "FedTherapist: Mental Health Monitoring with User-Generated Text Data via Federated Learning," in Proc. Conf. Empir. Methods Nat. Lang. Process., 2023, pp. 11282-11295.

\bibitem{uzuner2011i2b2} Ö. Uzuner, B. R. South, S. Shen, and S. L. DuVall, "2010 i2b2/VA challenge on concepts, assertions, and relations in clinical text," J. Am. Med. Inform. Assoc., vol. 18, no. 5, pp. 552–556, 2011.

\bibitem{johnson2016mimic} A. E. Johnson, T. J. Pollard, L. Shen, L. W. H. Lehman, M. Feng, M. Ghassemi, B. Moody, P. Szolovits, L. A. Celi, and R. G. Mark, "MIMIC-III, a freely accessible critical care database," Sci. Data, vol. 3, no. 1, pp. 1–9, 2016.

\bibitem{mullenbach2018explainable} J. Mullenbach, S. Wiegreffe, J. Duke, J. Sun, and J. Eisenstein, "Explainable prediction of medical codes from clinical text," in Proc. Conf. North Am. Chapter Assoc. Comput. Linguist.: Hum. Lang. Technol., 2018, pp. 1101–1111.

\bibitem{xing2019local} L. Xing, Y. Li, and T. Li, "Local concentrating, not shear stress, that may lead to possible instability of protein molecules during syringe injection: a fluid dynamic study with two-phase flow model," PDA J. Pharm. Sci. Technol., vol. 73, no. 3, pp. 260-275, 2019.

\bibitem{li2021physics} Y. Li, "Physics-based Simulation of Tablet Disintegration and Dissolution," Ph.D. dissertation, Purdue University, 2021.

\bibitem{li2019integrating} Y. Li and T. Li, "Integrating Lattice Boltzmann and Bonded Particle Models to Simulate Tablet Disintegration and Dissolution," in National Institute for Pharmaceutical Technology \& Education, 2019.

\bibitem{chen2024adaptive} J. Chen, B. Liu, X. Liao, J. Gao, H. Zheng, and Y. Li, "Adaptive Optimization for Enhanced Efficiency in Large-Scale Language Model Training," in Proc. Int. Conf. Front. Technol. Inf. Comput., 2024, pp. 1315-1319.

\bibitem{wang2025characteristics} S. Wang, S. Su, F. Cai, C. Zhou, M. Du, Y. Li, and X. Jiang, "Characteristics analysis of nitrogen gas fracturing in coal treated by liquid nitrogen freeze--thaw based on numerical model," J. Brazilian Soc. Mech. Sci. Eng., vol. 47, no. 2, pp. 1-12, 2025.

\bibitem{wang2025mechanical} S. Wang, S. Su, F. Cai, H. Li, P. Hou, T. Teng, M. Du, Y. Li, H. Li, and X. Jiang, "Mechanical degradation characteristics and permeability evolution law of coal under liquid nitrogen freeze--thaw cycles," Phys. Fluids, vol. 37, no. 2, 2025.
\end{thebibliography}
\end{document}